\documentclass[review]{elsarticle}

\usepackage{lineno,hyperref}

\journal{Journal of \LaTeX\ Templates}

\usepackage{amsmath}
\usepackage{booktabs}  
\usepackage{longtable}
\usepackage{multirow}
\usepackage{tikz}
\usepackage{amssymb}
\usepackage{footnote}
\usepackage{graphicx}  
\usepackage{epstopdf}
\usepackage[referable]{threeparttablex}

\begin{document}

	\begin{frontmatter}
		
		\title{A general framework for adaptive two-index fusion attribute weighted naive Bayes\tnoteref{mytitlenote}}
		

		\author[mymainaddress]{Xiaoliang Zhou}
		\author[mymainaddress]{Dongyang Wu}
		\author[mymainaddress]{Zitong You}
		\author[mymainaddress]{Li Zhang\corref{mycorrespondingauthor}}
		\cortext[mycorrespondingauthor]{Corresponding author}
		\ead{lizhang@njfu.edu.cn}
		
		\author[mymainaddress]{Ning Ye}
		
		\address[mymainaddress]{College of Information Science and Technology, Nanjing Forestry University, Nanjing, Jiangsu, 210037, China.}

		\begin{abstract}
			Naive Bayes(NB) is one of the essential algorithms in data mining. However, it is rarely used in reality because of the attribute independent assumption. Researchers have proposed many improved NB methods to alleviate this assumption. Among these methods, due to high efficiency and easy implementation, the filter attribute weighted NB methods receive great attentions. However, there still exists several challenges, such as the poor representation ability for single index and the fusion problem of two indexes. To overcome above challenges, we propose a general framework for Adaptive Two-index Fusion attribute weighted  NB(ATFNB). 
			Two types of data description category are used to represent the correlation between classes and attributes, intercorrelation between attributes and attributes, respectively. ATFNB can select any one index from each category. Then, we introduce a switching factor $\beta$ to fuse two indexes, which can adaptively adjust the optimal ratio of the two index on various datasets. And a quick algorithm is proposed to infer the optimal interval of switching factor $\beta$. Finally, the weight of each attribute is calculated using the optimal value $\beta$ and is integrated into NB classifier to improve the accuracy. The experimental results on 50 benchmark datasets and a Flavia dataset show that ATFNB outperforms the basic NB and state-of-the-art filter weighted NB models. In addition, the ATFNB framework can improve the existing two-index NB model by introducing the adaptive switching factor $\beta$. Auxiliary experimental results demonstrate the improved model significantly increases the accuracy compared to the original model without the adaptive switching factor $\beta$. 
		\end{abstract}
	\end{frontmatter}


	\section{Introduction}
	
	The Naive Bayes (NB) is a classical classification algorithm. Due to its simplicity and efficiency, it is widely used in many fields such as data mining and pattern recognition.
	
	Assume that a dataset $D = \{ x_1,x_2,…,x_m\}$ contains $m$ training instances, an instance $x_i$ can be represented by an $n$-dimensional attribute value vector $ < x_{i1},x_{i2},…,x_{in} >$. NB uses Equation \ref{key1} to predict the class label of the instance $x_i$.
	\begin{equation} \label{key1}
		c(x_i )=\mathop{argmax}_{c \in C} P(c) \prod_{j=1}^n P(x_{ij}|c)
	\end{equation}
	where $C$ is the set of all possible class labels $c$, $n$ is the number of attributes, and $x_{ij}$ represents the value of the $j$-$th$ attribute of the $i$-$th$ instance. $P(c)$ is the prior probability of class $c$, and $  P(x_{ij} | c) $ is the conditional probability of the attribute value $ x_{ij} $ given the class $c$, which can be calculated by Equation \ref{key2} and \ref{key3}, respectively.
	\begin{equation}\label{key2}
		P(c)=\frac{\sum_{i=1}^m {\delta(c_i,c)+1}} {m+\vartheta(C)}  
	\end{equation}
	\begin{equation}\label{key3}
		P(x_{ij}|c)=\frac{\sum_{i=1}^m {\delta(c_i,c)\delta(x_{ij},A_{j})+1}}{\sum_{i=1}^m {\delta(c_i,c)}+\vartheta(A_{j})}  
	\end{equation}
	where $A_j$ represents all the values of the $j$-$th$ attribute in training instances. $\vartheta$(·) is a custom function to calculate the number of unique data in $C$ or $A_j$. $c_i$ denotes the correct class label for the $i$-$th$ instance. $\delta$(·) is a binary function, which takes the value 1 if $c_i$ and c are identical and 0 otherwise \cite{zhang2020class}.
	
	Duo to the attribute independence assumption, NB is a simple, stable, easy to implement, and better classification algorithm for various applications. However, the real data is complicated and diverse, which is difficult to satisfy this assumption. Thus, researchers proposed many methods to reduce the influence of attribute independence assumption. These methods can be divided into six categories: structure extension is that directed arcs are modelled to represent the dependence relationship between attributes \cite{koivisto2004exact,friedman2003being,friedman1997bayesian,jiang2008novel,jiang2016structure}. Fine tuning is to adjust the probability value to find a good estimation of the desired probability term \cite{diab2017using,el2014fine}. The purpose of instance selection is to construct NB model on a subset of training set instead of the whole training set \cite{ryu2015hybrid,xie2002snnb}. Instance weighting is that instances are assigned different weights by different strategies \cite{zhang2021attribute,elkan1997boosting,jiang2010improving}. 
	Attribute selection is the process of removing redundant attributes \cite{chen2009feature,choubey2017classification,hall2003benchmarking,jiang2012not,lee2006information,deng2019feature}. Distinguished from attribute selection, attribute weighting assigns weight to each attribute in order to relax the independence assumption and make NB model more flexible.\cite{lee2018information,yu2019toward,jiang2019class,wu2015self,hall2006decision,taheri2014attribute,jiang2016deep,zhang2016two}. 
	
	In this paper, we focus our attention on attribute weighting, which is further divided into wrapper methods and filter methods. The wrapper methods optimize the weighted matrix by using gradient descent to improve classification performance. \textit{Wu et al}. proposed a weighted NB algorithm based on differential evolution, which gradually adjusts the weights of attributes through evolutionary algorithms to improve the prediction results\cite{wu2011attribute}. \textit{Zhang et al}. proposed two attribute value weighting models based on conditional log-likelihood and mean square error \cite{zhang2020class}. However, these methods are often less efficient due to the time-consuming optimization process. Another category obtain the weights by analyzing the correlation of attributes\cite{lee2018information,yu2019toward,hall2006decision,jiang2016deep,zhang2016two}. Since correlation can be easily and efficiently obtained by various measurement indexes, the computational efficiency of filter methods obviously increase. Related filter methods will be detailed introduced in Section 2. Although filter methods have some advantages such as flexible and computational efficient, there are still two problems. Most of methods utilize a single index, which expresses the data characteristic, to determine the attribute weight. However, a single index can not comprehensively discovery information of dataset. In order to fully dig up the information of dataset, two-index fusion method was proposed, which can achieve better performance \cite{jiang2018correlation}. However, the ratio of two indexes become the second problem. The method assumes that the contributions of two indexes are equivalent and ignores the difference in contribution between two indexes.
	
	To overcome the above problems, we propose a general framework for Adaptive Two-index Fusion attribute weighted Naive Bayes (ATFNB). ATFNB can select any index from two categories of data description, respectively. The first category describes the correlation between attributes and classes, and the second category describes the intercorrelation between attributes and attributes. Once two indexes are selected, ATFNB fuses two indexes by introducing a switching factor $\beta$. Due to the diversity of datasets, the switching factor $\beta$ can be adaptively to get the optimal ratio between two indexes. What is more, a quick algorithm is proposed to obtain the optimal value of switching factor $\beta$. To verify the effectiveness of ATFNB, we conduct extensive experiments on 50 UCI dataset and a Flavia dataset. Experimental results show that ATFNB has a better performance compared to NB and state-of-the-art filter NB models
	
	The rest of the paper consists of the following parts. Section 2 comprehensively reviews the filter attribute weighted methods. Section 3 proposes a general framework for adaptive two-index fusion attribute weighted naive Bayes. Section 4 presents the experimental datasets, setting and results. Section 5 further discusses the experimental results. Finally, Section 6 summarizes the research and gives the future work.

	\section{Related work}
	Given a dataset $D$ with $n$ attributes and $K$ classes. The naive Bayes weight matrix  is shown in Table \ref{T1}. 
	\begin{table}[htbp]\centering
		\caption{The naive Bayes weight matrix}
		\label{T1}
		\begin{tabular*}{1\textwidth}{@{\extracolsep{\fill}} ccccccc}
			\toprule
			&$A_1$&$A_2$&......&$A_{n-2}$&$A_{n-1}$&$A_n$\\
			\midrule
			$c_1$&$w_1$&$w_2$&......&$w_{n-2}$&$w_{n-1}$&$w_n$\\
			$c_2$&$w_1$&$w_2$&......&$w_{n-2}$&$w_{n-1}$&$w_n$\\
			$...$&$...$&$...$&......&$...$&$...$&$...$\\
			$c_{K-1}$&$w_1$&$w_2$&......&$w_{n-2}$&$w_{n-1}$&$w_n$\\
			$c_{K}$&$w_1$&$w_2$&......&$w_{n-2}$&$w_{n-1}$&$w_n$\\
			\bottomrule
		\end{tabular*}
	\end{table}
	The naive Bayes incorporates the attribute weight into the formula as follows:
	\begin{equation} \label{key4}
		\hat{c}(x_i)=\mathop{argmax}_{c \in C} P(c) \prod_{j=1}^n P(x_{ij}|c)^{w_j}
	\end{equation}
	where $w_j$ is the weight of the $j$-$th$ attribute $A_j$. The most critical issue of filter weighted NB methods is how to determine the weight $w_j$ of each attribute, which has attracted more great attention. Many weighted NB methods are proposed based on various measurements of attribute weighted. Here, we introduce several state-of-the-art filter weighted NB methods.
	
	\textit{Ferreira et al.} firstly proposed a weighted Naive Bayes to alleviate the independence assumption, which assigned weights to different attributes \cite{ferreira2001weighted}. Based on this idea, \textit{Zhang et al.} presented an attribute weighted model based on gain ratio (WNB)\cite{zhang2004learning}. Attribute with higher gain ratio deserved higher weight in WNB. Therefore, the weight of each attribute can be defined by Equation \ref{key5}. 
	\begin{equation} \label{key5}
		w_j=\frac{GainRatio(D,A_j)}{\frac{1}{n}\sum_{j=1}^nGainRatio(D,A_j)}
	\end{equation}
	where $ GainRatio(D,A_j) $ is the gain ratio of attribute $A_j$ \cite{quinlan1993c4} . 
	
	Then, \textit{Lee et al.}  proposed a novel model that used the Kullback-Leibler metric to calculate the weight of each attribute \cite{lee2011calculating}. This model was certain information between each attribute and the corresponding class label $c$, which was obtained by Kullback-Leibler \cite{kullback1951information} measuring the difference between the prior distribution and the posterior distribution of the target attributes. The weight value of the $j$-$th$ attribute is shown in Equation \ref{key6}.
	\begin{equation}\label{key6}
		w_j=\frac{1}{Z}\frac{\sum_{j|i}P(x_{ij})KL(c|x_{ij})}{-\sum_{j|i}P(x_{ij})log(P(x_{ij}))}
	\end{equation}
	where $P(x_{ij})$ means the probability of the value $x_{ij}$, and $ Z=\frac {\sum_i^n w_i}{n} $ is a normalization constant. $KL(c|x_{ij})$ is the average mutual information between the class label $c$ and the attribute value of $x_{ij}$.
	
	Next, \textit{Jiang} team proposed a series of filter attribute weight methods, which included Deep Feature Weighting (DFW) \cite{jiang2016deep} and Correlation-based Feature Weighting (CFW) \cite{jiang2018correlation}. DFW assumed that more independent features should be assigned higher weight. The correlation-based feature selection was used to evaluate the degree of dependence between attributes\cite{hall2000correlation}. According to this selection, the best subset was selected from the attribute space. The weight value assigned to the selected attribute was 2, and the weight value assigned to other attributes was 1, as shown in Equation \ref{key7}.
	
	\begin{equation}\label{key7}
		w_j =
		\left\{
		\begin{array}{cc}
			2,  &if \ A_j \ is\ selected.  \\
			1,  &otherwise.
		\end{array}
		\right.
	\end{equation}
	
	Compared with the above methods with a single index, CFW was the first two-index weighted NB method, which used the attribute-class correlation and the average attribute-attribute intercorrelation to constitute the weight of each attribute. The mutual information was measured the attribute-class correlation and the attribute-attribute intercorrelation, defined as Equation \ref{key8} and \ref{key9}, respectively.
	\begin{equation}\label{key8}
		I(A_j;C)=\sum_{a_j}\sum_c P(a_j,c) log \frac{P(a_j,c)}{P(a_j)P(c)}
	\end{equation}
	\begin{equation}\label{key9}
		I(A_i;A_j)=\sum_{a_i}\sum_{a_j} P(a_i,a_j) log \frac{P(a_i,a_j)}{P(a_i)P(a_j)}
	\end{equation}
	where $a_i$ and $a_j$ represent the values of attributes $A_i$ and $A_j$ respectively. $I(A_j;C)$ is the correlation between the attribute $A_j$ and class C. $I(A_i;A_j)$ is the redundancy between two different attributes $A_i$ and $A_j$. Finally, the weight of the attribute $w_j$ is defined as Equation \ref{key13}.
	
	\begin{equation}\label{key13}
		w_j=\frac{1}{1+e^{-({NI(A_j;C)-\frac{1}{n-1}\sum_{{j=1} \cap {j \neq i}}^n NI(A_i;A_j)})}}
	\end{equation}
	where $NI(A_j;C)$ and $NI(A_i;A_j )$ are the normalized values, which respectively represent the maximum correlation and the maximum redundancy.

	\section{ATFNB}
	\subsection{The general framework of ATFNB}
\begin{figure}[h]\centering
		\includegraphics[width=0.7\textwidth]{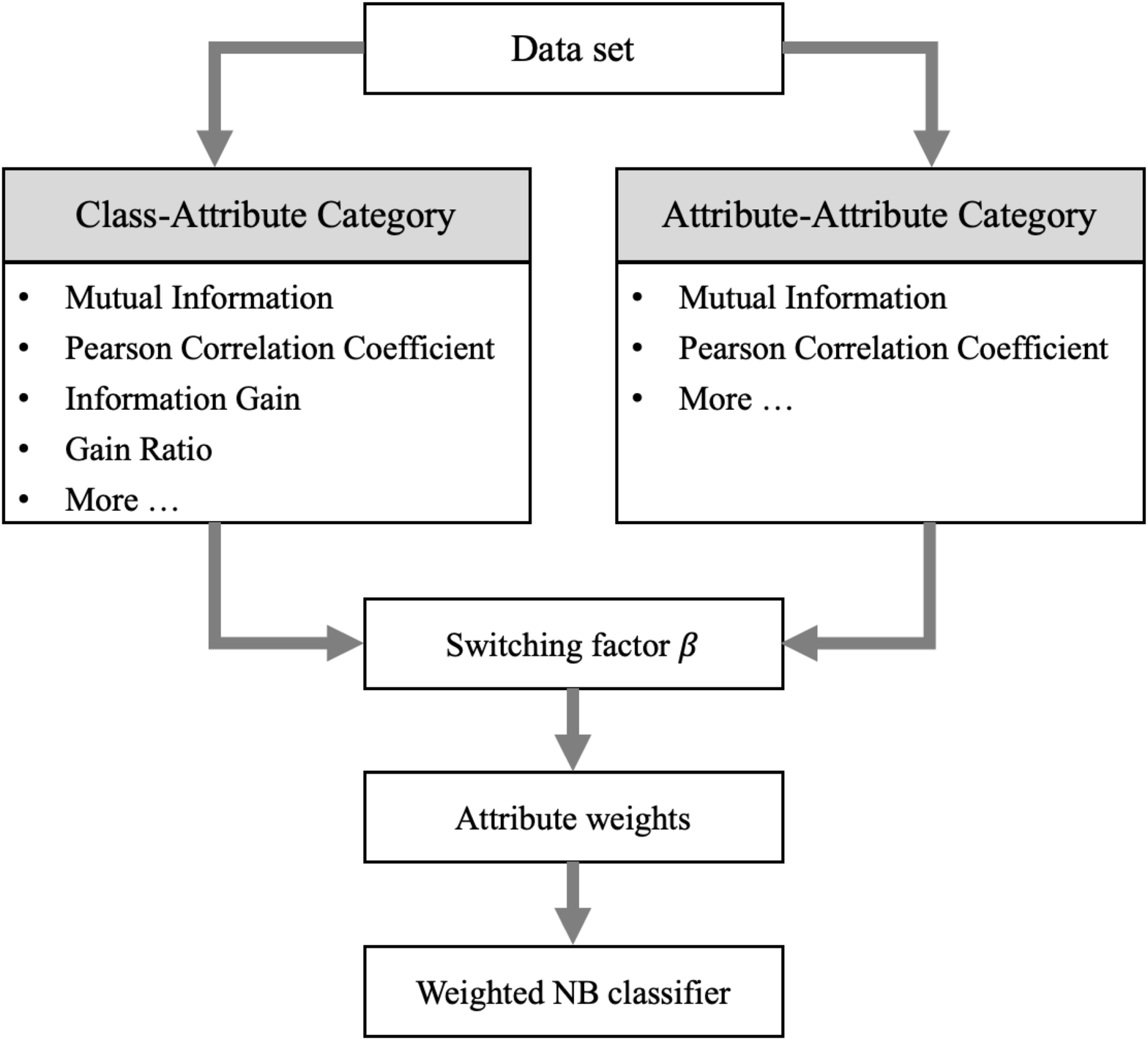}
		\caption{A general framework for adaptive two-index fusion attribute weighted NB }
		\label{F3.1}
	\end{figure}

	The filter weighted NB methods assign a specific weight for each attribute to alleviate the independence assumption. However, there are still some challenges, such as the poor representation ability for single index and the fusion problem of two indexes. Therefore, we propose a general framework for adaptive two-index fusion attribute weighted NB. The framework of ATFNB is shown in Figure \ref{F3.1}. Given a dataset, two indexes are selected from class-attribute category and attribute-attribute category, respectively. Then, the switching factor $\beta$ is utilized to fuse the two indexes, and adaptively generate the optimal ratio value. Next, the weight of each attribute is calculated via the optimal switching factor $\beta$. Finally, the attribute weights are incorporated into the NB classifier to predict the class labels.
	
	\subsection{Index Selection}
	As shown in Figure \ref{F3.1}, the ATFNB framework contains two widely used types of the attribute correlation: class-attribute and attribute-attribute. The class-attribute category is to measure the correlation between attributes and classes. The stronger the correlation between attribute and class, the more significant the attribute's contribution to the classification. Thus, the index value is positively correlated with the weight. Common indexes in this category contain mutual information, Pearson correlation coefficient, information gain, and gain ratio, etc. The attribute-attribute category is to measure the redundancy between attributes. In order to satisfy the independence assumption of Naive Bayes as much as possible, attributes with high redundancy are assigned small weights. Thus, the weight is inversely correlated to the index value. Common measures of redundancy between attributes include mutual information and Pearson correlation coefficient, etc.
	
	By selecting different indexes, the ATFNB framework can become any weighted NB model, including the existing weighted NB models. If only the gain ratio
	 is selected from class-attribute category, ATFNB will degenerate into the single-index WNB model. If the class-attribute and attribute-attribute category both choose the mutual information, ATFNB will become the two-index CFW model. Thus, the index selection is a critical step in the ATFNB framework. Any two indexes are selected from the two categories can generate various models, which may achieve different results. In Section 5.3, the classification performances of different index selections are detailed discussed. 
	
	\subsection{A quick algorithm for the switching factor $\beta$}
	
	Since the significantly discriminative attribute should be highly correlated with the class and has low redundancy with other attributes, its weights should be positively associated with the difference between class-attribute correlation and attribute-attribute intercorrelation\cite{jiang2018correlation}. The mathematics formula of the weight $w_j$ can be defined by Equation \ref{key14}.  
	
	\begin{equation}\label{key14}
		w_j=\underbrace{class\_attribute}_{CA_j}-\underbrace{attribute\_attribute}_{AA_j}
	\end{equation}
	where $CA_j$ and $AA_j$ represents the values of the selected class-attribute index and attribute-attribute index, respectively. The existing two-index methods, CFW, consider that the contributions of two indexes are equivalent\cite{jiang2018correlation}. However, various indexes contain different characteristic, and the equivalent contribution of two indexes is unreasonable. Thus, we introduce a switching factor $\beta$ to adaptively control the ratio of two indexes. After incorporating the switching factor $\beta$, Equation \ref{key14} can be rewritten as Equation \ref{key15}.
	\begin{equation}\label{key15}
		w_j=\beta\times {CA_j}-(1-\beta)\times {AA_j}
	\end{equation}
	where the switching factor $\beta \in [0,1]$ . 
	
	Conventionally, the step-length searching strategy can be applied to search the optimal interval of $\beta$. But the accuracy and computational efficiency are effected by the step size. When the step size gets smaller, the optimal interval of $\beta$ is more accurate but get very slower.  Thus, we propose a quickly algorithm to calculate the optimal interval of switching factor $\beta$. Firstly, the basic weighted NB model (Equation \ref{key4}) is logarithmically transformed, and the detailed transformation process is shown as follow. 
	
	\begin{equation}\label{key16}
		\begin{aligned}
			T(x_i,c)&=log(P(c))+\sum_{j=1}^n log{(P(x_{ij}|c))}^{w_j} \\
			&=log(P(c))+\begin{bmatrix} w_1,...,w_j \end{bmatrix} \begin{bmatrix} log(P(x_{i1}|c)) \\ ... \\ log(P(x_{ij}|c)) \end{bmatrix}\\
			&=\beta\underbrace{\begin{bmatrix} (CA_1+AA_1),...,(CA_j+AA_j) \end{bmatrix} \begin{bmatrix} log(P(x_{i1}|c)) \\ ... \\ log(P(x_{ij}|c))\end{bmatrix} }_{K_c} \\
			&+ \underbrace{log(P(c))-\begin{bmatrix} P_1,...,P_j \end{bmatrix} \begin{bmatrix}  log(P(x_{i1}|c)) \\ ... \\ log(P(x_{ij}|c))  \end{bmatrix}}_{M_c}\\
			&=\beta_i\times K_c +M_c
		\end{aligned}
	\end{equation}
	where $T(x_i,c)$ is the probability value that the instance $x_i$ belonging to class c. $\beta_i$ is the interval when instance $x_i$ is correctly classified. 
	
	Based on Equation \ref{key16}, a probability set $S_i$ can be constructed to store the probability values of instances $x_i$ belonging to different classes.
	The probability set of $x_i$ is $S_i=\{T(x_i,c_1 ),T(x_i,c_2 ),…,T(x_i,c_K)\}$. If the correct label of $x_i$ is $c_k$, $T(x_i,c_k)$ should be greater than the other probability values in $S_i$.  This can be defined as follow.
	\begin{equation}\label{key18}
		T(x_i,c_k) \textgreater \{S_i-T(x_i,c_k)\}  
	\end{equation}

	When the instance $x_i$ is correctly classified, the interval of $\beta_i$ can be obtained. For $m$ instances, a set $G=\{\beta_1,\beta_2,…, \beta_m\}$ contains the $m$ interval corresponding to each instance. To calculate the optimal interval $\beta^*$ from G, that any value in the interval can obtain the same classification accuracy on the training set. The upper and lower bounds of all intervals in $G$ are sorted in ascending order $Q=\{value_1,value_2,…, value_{q}\}$. Any two adjacent values in Q are regarded as the lower and upper bounds of a subinterval. Thus, Q can generate $q-1$ subintervals. The subintervals in $R=\{\gamma_1,\gamma_2,...,\gamma_{q-1}\}$ satisfying Equation \ref{key19} are taken as $\beta^*$.
	\begin{equation}\label{key19}
		max\{\sum_{i=1}^m \tau(\gamma_1,\beta_i),\sum_{i=1}^m \tau(\gamma_2,\beta_i),...,\sum_{i=1}^m \tau(\gamma_{q-1},\beta_i) \}
	\end{equation}
	where $\tau$(·) is a binary function, which takes the value 1 if $\gamma_{q-1}$ is a subset of $\beta_i$ and 0 otherwise, as shown in Equation \ref{key20}.
	
	\begin{equation}\label{key20}
		\tau =
		\left\{
		\begin{array}{cc}
			1,  &if  \gamma_{q-1} \subseteq \beta_i.  \\
			0,  &otherwise.
		\end{array}
		\right.
	\end{equation}

According to the above derivation processes, the Quick algorithm for the Switching Factor(QSF) is described in Algorithm 1.\\
	
	\begin{tabular*}{1\textwidth}{@{\extracolsep{\fill}} l}
		\toprule
		Algorithm 1${\colon}$QSF\\
		\midrule
		\textbf{Input$\colon$class-attribute($CA_j$)}, \textbf{attribute-attribute($AA_j$)}, \textbf{Dataset D}  \\
		For each instance $x_i$ in D$\colon$ \\
		\quad\quad For each class c in C$\colon$\\
		\quad\quad\quad\quad Calculate $K_c$ and $M_c$ in Equation \ref{key16}.\\
		\quad\quad\quad\quad According $K_c$ and $M_c$, get $T(x_i,c)$.\\
		\quad\quad End\\
		\quad\quad $S_i=\{T(x_i,c_1 ),T(x_i,c_2 ),…,T(x_i,c_K)\}$\\
		\quad\quad If instance $x_i$ label is $c_k$$\colon$\\
		\quad\quad\quad\quad You can find value that satisfies Equation \ref{key18}, it is recorded as $\beta_i$, \\
		\quad\quad\quad\quad otherwise $\beta_i$=$\emptyset$.\\
		\quad\quad End\\
		\quad\quad $G=\{\beta_1,\beta_2,…, \beta_m\}$ \\
		End\\
		For each $\gamma_{q-1}$\\
		\quad\quad Find the subinterval $\beta^*$  that conforms to Equation \ref{key19}.\\
		End\\
		\textbf{Output$\colon$$\beta^*$}\\
		\bottomrule
	\end{tabular*}\\

Any value in $\beta^*$ can achieve consistent classification accuracy in dataset $D$, so we choose any value from optimal interval $\beta^*$. Once obtaining the value of switching factor, the weight $w_i$ can be calculated by Equation  \ref{key15}.

	\subsection{The Implementation of ATFNB}
	
	The general framework of ATFNB is briefly described in Algorithm 2. According to Algorithm 2, we can see that how to select two indexes $AA_j$ and $CA_j$, and how to learn the switching factor $\beta$ are two crucial problems. To select $AA_j$ and $CA_j$, several indexes are listed in Subsection 3.2. To learn the value of the switching factor $\beta$, we single out a QSF algorithm in Subsection 3.3. Once the value of the switching factor $\beta$ is obtained, we can use Equation \ref{key15} to calculate the weights of each attributes. Finally, these weights are applied to construct an attribute weighted NB classifier. \\
	\begin{tabular*}{1\textwidth}{@{\extracolsep{\fill}} l}
		\toprule
		Algorithm 2${\colon}$ATFNB Framework\\
		\midrule
		\textbf{Input$\colon$} Training set \textbf{D}, Test set\textbf{ X} \\
		\quad\quad(1)\quad For each attribute $A_j$ in D \\
		\quad\quad\quad\quad\quad\quad calculating \textbf{(attribute-attribute) index $AA_j$}\\
		\quad\quad\quad\quad\quad\quad calculating \textbf{(class-attribute) index $CA_j$}.\\
		\quad\quad(2)\quad 	According to \textbf{QSF}, the value of the \textbf{switching factor $\beta$} is solved. \\
		\quad\quad(3)\quad 	According to \textbf{Equation \ref{key15}}, weight matrix is obtained. \\
		\quad\quad(4)\quad 	According to \textbf{Equation \ref{key4}}, 	the class label of each instance in X is predicted. \\
		\textbf{Output$\colon$Class label of instances in X}\\
		\bottomrule
	\end{tabular*}

	\section{Experiments and Results}
	\subsection{Experimental data}
	To verify the effectiveness of ATFNB, a collection of 50 benchmark datasets and 15 groups of leaf dataset are conducted.
	
	The 50 benchmark classification datasets are chosen from the University of California at Irvin (UCI) repository\cite{asuncion2007uci},which represent various fields and data characteristics listed in Table \ref{T2}. We use the mean of the corresponding attribute to replace the missing data values in each dataset, then apply chi-square-based algorithm to discretize the numerical attribute values \cite{kerber1992chimerge}. The amount of discretization of each attribute is consistent with the number of types of class labels.\\
	
	\begin{ThreePartTable}

		\begin{longtable}{lccc}
			\caption{Descriptions of 50 UCI datasets used in the experiments}
			\label{T2}\\
			\toprule
			Dataset&Instance number&Attribute number&Class number \\
			\midrule
			\endhead
			\bottomrule
			\hline
			\multicolumn{4}{r}{\textit{Continued on next page}} \\
			\endfoot
			\bottomrule
			
			\endlastfoot
			\emph{abalone}&4177&8&3\\
			\emph{acute}&120&6&2\\
			\emph{aggregation}&788&2&7\\
			\emph{balance-scale}&625&4&3\\
			\emph{bank}&4521&16&2\\
			\emph{banknote}&1372&4&2\\
			\emph{blood}&748&4&2\\
			\emph{breast-cancer}&286&9&2\\
			\emph{breast-tissue}&106&9&6\\
			\emph{bupa}&345&6&2\\
			\emph{car}&1728&6&4\\
			\emph{chart\_Input}&600&60&6\\
			\emph{climate-simulation}&540&18&2\\
			\emph{congressional-voting}&435&16&2\\
			\emph{connectionist}&208&60&2\\
			\emph{dermatology}&366&34&6\\
			\emph{diabetes}&768&8&2\\
			\emph{ecoli}&336&7&8\\
			\emph{energy-y1}&768&8&3\\
			\emph{fertility}&100&9&2\\
			\emph{glass}&214&9&6\\
			\emph{haberman-survival}&306&3&2\\
			\emph{iris}&150&4&3\\
			\emph{jain}&373&2&2\\
			\emph{knowledge}&172&5&4\\
			\emph{libras}&360&90&15\\
			\emph{low-res-spect}&531&100&9\\
			\emph{lymphography}&148&18&4\\
			\emph{magic}&19020&10&2\\
			\emph{mammographic}&961&5&2\\
			\emph{promoters}&106&57&2\\
			\emph{splice}&3190&60&3\\
			\emph{nursery}&12960&8&5\\
			\emph{page-blocks}&5473&10&5\\
			\emph{pima}&768&8&2\\
			\emph{planning}&182&12&2\\
			\emph{post-operative}&90&8&3\\
			\emph{robotnavigation}&5456&24&4\\
			\emph{seeds}&210&7&3\\
			\emph{sonar}&208&60&2\\
			\emph{soybean}&683&35&18\\
			\emph{spect}&265&22&2\\
			\emph{synthetic-control}&600&60&6\\
			\emph{tic-tac-toe}&958&9&2\\
			\emph{titanic}&2201&3&2\\
			\emph{twonorm}&7400&20&2\\
			\emph{wall-following}&5456&24&4\\
			\emph{waveform}&5000&21&3\\
			\emph{wilt}&4839&5&2\\
			\emph{wine}&178&13&3\\	
		\end{longtable}
	\end{ThreePartTable}
	
	The Flavia dataset contains 32 types of leaf and each leaf has 55-77 pieces. Four texture and ten shape features of each leaf are extracted based on the grayscale and binary images\cite{sachar2021survey}. We construct 15 groups to comparative experiments, and each group randomly selects 15 kinds of leaves from the whole Flavia dataset. The detailed characteristics of 15 groups are listed in Table \ref{T3}. Then the same pro-processing pipeline as the UCI dataset are applied to discretize continuous attributes.
	\begin{table}[h]\centering
		\caption{Descriptions of 15 groups from Flavia dataset used in the experiments}
		\label{T3}
		\begin{tabular*}{1\textwidth}{@{\extracolsep{\fill}} lccc}
			\toprule
			Group&Instance number&Attribute number&Class number \\
			\midrule
			\emph{$G\_1$}&869&14&15\\
			\emph{$G\_2$}&888&14&15\\
			\emph{$G\_3$}&865&14&15\\
			\emph{$G\_4$}&887&14&15\\
			\emph{$G\_5$}&884&14&15\\
			\emph{$G\_6$}&919&14&15\\
			\emph{$G\_7$}&892&14&15\\
			\emph{$G\_8$}&881&14&15\\
			\emph{$G\_9$}&864&14&15\\
			\emph{$G\_10$}&879&14&15\\
			\emph{$G\_11$}&895&14&15\\
			\emph{$G\_12$}&888&14&15\\
			\emph{$G\_13$}&927&14&15\\
			\emph{$G\_14$}&924&14&15\\
			\emph{$G\_15$}&904&14&15\\
			\bottomrule
		\end{tabular*}
	\end{table}

	\subsection{Experimental Setting}
	ATFNB is a general framework of attribute weighted naive Bayes, which can adaptively fuse any two indexes. According to Figure \ref{F3.1}, we select two simple and popular indexes from two categories: information gain from class-attribute category and Pearson correlation coefficient from attribute-attribute category. Notably, ATFNB refers to a specific NB model fused the above two indexes in the following experiments, and no longer represents a general framework.
	
	For the class-attribute category, information gain describes the information content provided by the attribute for the classification. The formula of information gain is shown in Equation \ref{key22}.
	
	\begin{equation}\label{key22}	
		Gain(D;A_j)=Ent(D)-\sum_{v=1}^{V}\frac{|D^v|}{|D|} Ent(D^v)
	\end{equation}
	where $Ent(D)$ is the information entropy. The discrete attribute $A_j$ has V values  $\{a^1,a^2,…,a^V\}$. $D^v$ indicates that the $v$-$th$ branch node contains all the instances in dataset $D$, whose value is $a^V$ on the attribute $A_j$.
	
	For the attribute-attribute category, Pearson correlation coefficient is used to calculate the correlation between attributes $A_i$ and $A_j$, and the formula can be written as Equation \ref{key23}.
	\begin{equation}\label{key23}
		\rho(A_i;A_j)=|\frac{cov(A_i,A_j)}{\sigma_{A_i}\sigma_{A_j}}|
	\end{equation}
	where $cov(A_i,A_j)$ represents the covariance between attributes $A_i$ and $A_j$, $\sigma_{A_i} $ and $ \sigma_{A_j}$ represent the standard deviation of  $A_i$ and $A_j$, respectively.
	
	Once obtaining above two indexes, the weight of each attribute $A_j$ can be calculated as Equation \ref{key24}.
	\begin{equation}\label{key24}
		w_j=\beta \times NGain(D;A_j)-(1-\beta)\times avg\_PCC(A_j)
	\end{equation}
	where $NGain(D;A_j)$ is expressed as the normalized value of attribute information gain, and  $avg\_PCC(A_j)$ represents the average degree of redundancy between the $i$-$th$ attribute and other attributes. The formula of $avg\_PCC(A_j)$  is shown in Equation \ref{key25}.
	\begin{equation}\label{key25}
		avg\_PCC(A_j)=\frac{1}{n-1}\sum_{{j=1}\cap{j\neq i}}^nN\rho(A_i;A_j)
	\end{equation}
	where $N\rho(A_i;A_j)$ is expressed as the normalized value between attributes $A_i$ and $A_j$.
	
	To validate the classification performance, we compare ATFNB to standard NB and two existing state-of-the-art filter weighted methods. In addition, the original CFW is a specific model of our framework under the switching factor $\beta$=0.5. When the switching factor $\beta$ of CFW can be adaptively obtained from the dataset, the original CFW evolves into CFW-$\beta$. Now, we introduce these comparisons and their abbreviations as follows:
	
	\begin{itemize}
		\item  NB: the standard naive Bayes model \cite{langley1992analysis}.
		\item  WNB: NB with gain ratio attribute weighting \cite{zhang2004learning}.
		\item  CFW: NB with MI class-specific and attribute-specific attribute weighting \cite{jiang2018correlation}.
		\item  CFW-$\beta$: CFW with the adaptive switching factor $\beta$.
	\end{itemize}

	\subsection{The effectiveness of the switching factor $\beta$}
	
\begin{figure}[!htbp]\centering
		\includegraphics[width=0.95\textwidth]{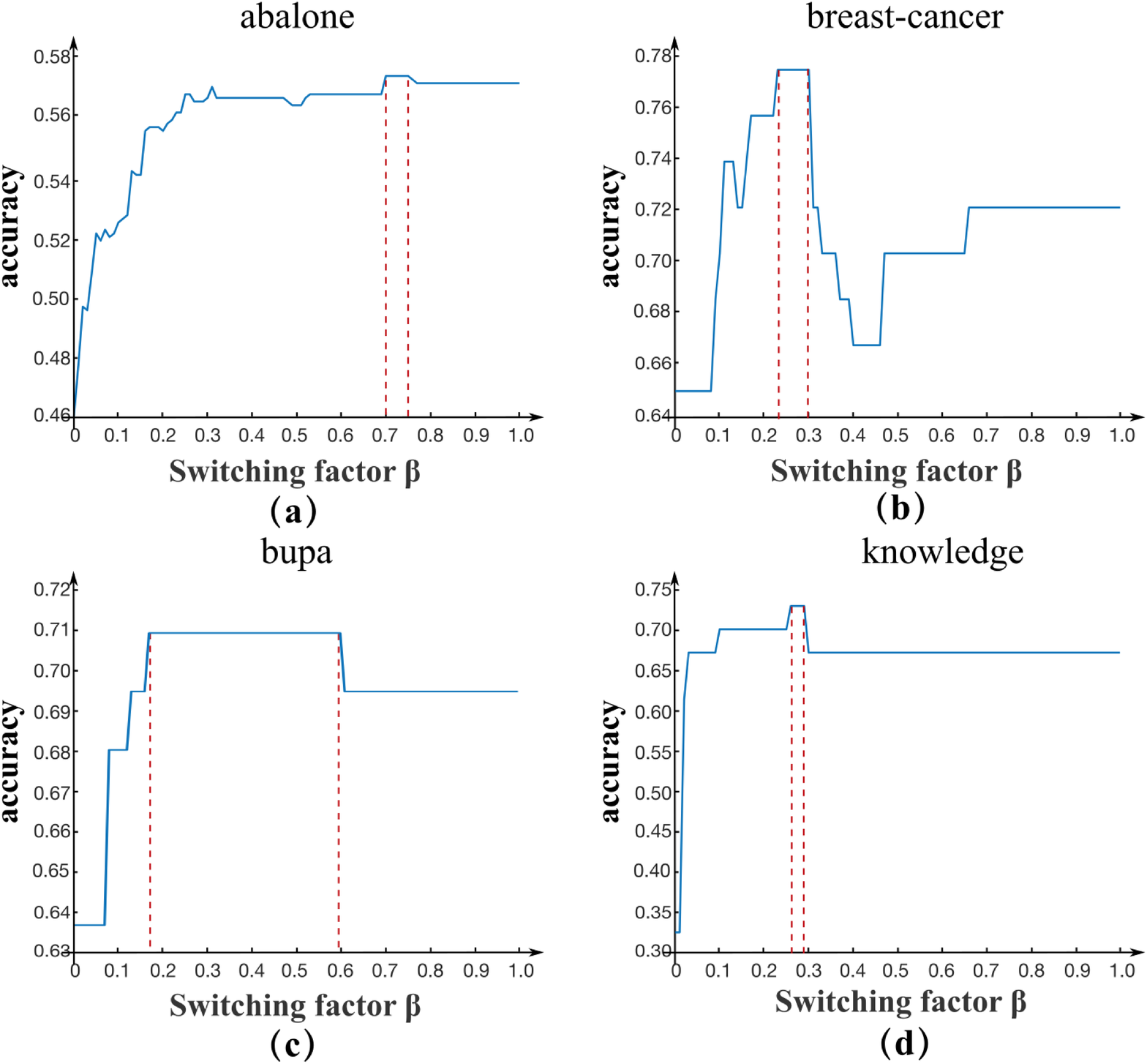}
		\caption{ The optimal switching factor $\beta$ of QSF and SLS algorithms in four datasets.The blue solid line represents the accuracy of each step size by SLS, and the red dotted lines represent the optimal interval obtained by QSF}
		\label{F4.3}
	\end{figure}

	The switching factor $\beta$ can be adaptively adjusted to obtain the optimal ratio for different datasets. In order to verify the effectiveness and efficiency of the switching factor $\beta$, we compare QSF algorithm with Step-Length Searching(SLS) algorithm. SLS algorithm generates $\beta$ with 0.01 as the step size. The optimal interval of switching factor $\beta$ by QSF and SLS algorithm in four datasets are shown in Figure \ref{F4.3}.

	From Figure \ref{F4.3}, either SLS or QSF algorithms, the optimal interval of $\beta$ in each dataset are biased. On \textit{abalone}, the lower bound of the interval of $\beta$ is greater than 0.5. On \textit{breast-cancer} and \textit{knowledge},
	the upper bound of the interval of $\beta$ is less than 0.5. Only the interval of $\beta$ in \textit{bupa} contains 0.5. Thus, it can be concluded that the switching factor $\beta$ value set as 0.5 is unreasonable for all datasets. In addition, it can be clearly seen that the interval size of switching factor $\beta$ is inconsistent. On \textit{bupa}, the interval size of $\beta$ is largest. On the contrary, the size is the smallest on \textit{knowledge}.

	\begin{table}[!htbp]\centering
		\caption{The optimal interval of switching factor $\beta$ and run-time by SLS and QSF}
		\label{T4}
		\begin{tabular*}{1\textwidth}{@{\extracolsep{\fill}} lcc|ccc}
			\hline
			\multicolumn{1}{c}{\multirow{2}{*}{Dataset}} & \multicolumn{2}{c|}{The interval of switching factor $\beta$} & \multicolumn{3}{c}{Time(s)} \\ \cline{2-6} 
			\multicolumn{1}{c}{}                                  & SLS                       & QSF                          & SLS      & QSF     & Speed  \\ \hline
			\textit{bupa}                                         & {[}0.17, 0.59{]}         & {[}0.1687, 0.5937{]}        & 7.4908   & 0.0119  & $\times$629    \\
			\textit{abalone}                                      & {[}0.70, 0.75{]}         & {[}0.6988, 0.7521{]}        & 16.826   & 0.1068  & $\times$157    \\
			\textit{breast-cancer}                                & {[}0.23, 0.31{]}         & {[}0.2257, 0.3129{]}        & 6.8023   & 0.0389  & $\times$174    \\
			\textit{knowledge}                                    & {[}0.27, 0.29{]}         & {[}0.2688, 0.2954{]}        & 6.1298   & 0.0229  & $\times$267     \\ \hline
		\end{tabular*}
	\end{table}
	
	The optimal interval of switching factor $\beta$ and run-time calculated by SLS and QSF are shown in Table \ref{T4}.
	From Table \ref{T4}, we can see that two optimal intervals obtained by SLS and QSF algorithms have a high coincidence degree. If we reduce the step size of SLS, the coincidence degree between two algorithms will further improve. Yet, SLS will become very inefficient. For QSF algorithm, the run-time is obviously faster than SLS, and speeds up 150 times at least. Therefore, QSF is not only more accurate than SLS, but also more efficient.
	
	\subsection{Experimental results on UCI Dataset}
	Table \ref{T5} shows the detailed classification accuracy results of five algorithms. All classification accuracy is obtained by averaging the results of 30 independent runs. Five algorithms are performed on the same training set and testing set. We conduct a group of experiments on 50 UCI dataset to compare ATFNB with NB, WNB, CFW and CFW-$\beta$ in terms of classification accuracy.

	\begin{ThreePartTable}
		\begin{TableNotes}
			\item[*]\label{tn:a}  indicates that ATFNB is significantly better than its competitors (NB, WNB, CFW) through two-tailed t-test at the $p$=0.05 significance level \cite{nadeau2003inference}. At the bottom of the table, \textit{G} represents the number of data sets with the highest classification accuracy among the four algorithms (ATFNB, WNB, CFW, NB). \textit{W} represents the classification accuracy is higher than ATFNB for the number of datasets, \textit{L} means the opposite of \textit{W}.	
		\end{TableNotes}
		
		\begin{longtable}{llllll}
			\caption{Classification accuracy comparisons for ATFNB versus NB, WNB, CFW, CFW-$\beta$ on UCI dataset}
			\label{T5}\\
			\toprule
			\textbf{Dataset}              & \textbf{NB} & \textbf{WNB} & \textbf{CFW} & \textbf{ATFNB} & \textbf{CFW-$\beta$} \\ 
			\midrule
			\endhead
			
			\hline
			\multicolumn{6}{r}{\textit{Continued on next page}} \\
			\endfoot
			\bottomrule
			\insertTableNotes \\
			\endlastfoot
			\textit{abalone}              & 0.5886      & 0.5871 *     & 0.5890 *     & 0.5908         & 0.5926          \\
			\textit{acute}                & 0.9958      & 0.9521 *     & 0.9948       & 0.9635         & 0.9813          \\
			\textit{aggregation}          & 0.9890      & 0.9882       & 0.9761 *     & 0.9875         & 0.9824          \\
			\textit{balance-scale}        & 0.8592 *    & 0.8728       & 0.8312 *     & 0.8984         & 0.8581          \\
			\textit{bank}                 & 0.8765      & 0.8831       & 0.8901       & 0.8822         & 0.9076          \\
			\textit{banknote}             & 0.8636      & 0.8468       & 0.8491       & 0.8498         & 0.8338          \\
			\textit{blood}                & 0.7597 *    & 0.7733       & 0.7720 *     & 0.7847         & 0.7990          \\
			\textit{breast-cancer}        & 0.7214 *    & 0.7059 *     & 0.7331       & 0.7472         & 0.7422          \\
			\textit{breast-tissue}        & 0.5727      & 0.5955       & 0.5818 *     & 0.6091         & 0.6158          \\
			\textit{bupa}                 & 0.6232      & 0.5942 *     & 0.6174 *     & 0.6333         & 0.6299          \\
			\textit{car}                  & 0.8523      & 0.6965 *     & 0.7671 *     & 0.8014         & 0.8101          \\
			\textit{chart\_Input}         & 0.9533      & 0.9367       & 0.9558       & 0.9455         & 0.9488          \\
			\textit{climate-simulation}   & 0.9137      & 0.9178       & 0.9174       & 0.9181         & 0.9209          \\
			\textit{congressional-voting} & 0.6149 *    & 0.6345 *     & 0.6253 *     & 0.6506         & 0.6614          \\
			\textit{connectionist}        & 0.7238 *    & 0.7429 *     & 0.7214 *     & 0.7667         & 0.7560          \\
			\textit{dermatology}          & 0.9797      & 0.9644       & 0.9757       & 0.9649         & 0.9665          \\
			\textit{diabetes}             & 0.7377      & 0.6584 *     & 0.7403       & 0.7422         & 0.7611          \\
			\textit{Ecoli}                & 0.8135      & 0.7706 *     & 0.7588 *     & 0.8245         & 0.7981          \\
			\textit{energy-y1}            & 0.8874      & 0.8225 *     & 0.8701       & 0.8463         & 0.8813          \\
			\textit{fertility}            & 0.8400 *    & 0.8500       & 0.8350 *     & 0.8650         & 0.8669          \\
			\textit{glass}                & 0.7023      & 0.6837 *     & 0.6930       & 0.7193         & 0.7233          \\
			\textit{haberman-survival}    & 0.7532 *    & 0.7468 *     & 0.7403 *     & 0.7710         & 0.7791          \\
			\textit{Iris}                 & 0.9133      & 0.9100 *     & 0.9167       & 0.9367         & 0.9099          \\
			\textit{Jain}                 & 0.9464      & 0.9368       & 0.9379       & 0.9397         & 0.9399          \\
			\textit{knowledge}            & 0.7371*     & 0.7743 *     & 0.7629 *     & 0.8057         & 0.7989          \\
			\textit{libras}               & 0.5903      & 0.5917       & 0.5847       & 0.5965         & 0.6122          \\
			\textit{low-res-spect}        & 0.8037 *    & 0.8018 *     & 0.8131*      & 0.8318         & 0.8411          \\
			\textit{lymphography}         & 0.8122 *    & 0.7889 *     & 0.8233       & 0.8334         & 0.8399          \\
			\textit{magic}                & 0.7300      & 0.6885 *     & 0.7411       & 0.7674         & 0.7782          \\
			\textit{mammographic}         & 0.8290 *    & 0.8394       & 0.8446       & 0.8549         & 0.8679          \\
			\textit{promoters}            & 0.9091 *    & 0.9045 *     & 0.9242 *     & 0.9545         & 0.9302          \\
			\textit{splice}               & 0.9475      & 0.9376 *     & 0.9580       & 0.9414         & 0.9677          \\
			\textit{nursery}              & 0.9043      & 0.8089 *     & 0.8812       & 0.8961         & 0.9002          \\
			\textit{page-blocks}          & 0.9300 *    & 0.9404       & 0.9545 *     & 0.9684         & 0.9690          \\
			\textit{pima}                 & 0.7338 *    & 0.6688 *     & 0.7330 *     & 0.7599         & 0.7613          \\
			\textit{planning}             & 0.6000 *    & 0.7189       & 0.6919 *     & 0.7378         & 0.7500          \\
			\textit{post-operative}       & 0.7222 *    & 0.8519 *     & 0.7593 *     & 0.9074         & 0.8489          \\
			\textit{robotnavigation}      & 0.8760 *    & 0.9159       & 0.9095       & 0.9179         & 0.9199          \\
			\textit{seeds}                & 0.8747      & 0.8622       & 0.8762       & 0.8655         & 0.8881          \\
			\textit{sonar}                & 0.7625      & 0.7429 *     & 0.7571 *     & 0.7734         & 0.7662          \\
			\textit{soybean}              & 0.9036      & 0.8730       & 0.9117       & 0.8781         & 0.9049          \\
			\textit{spect}                & 0.6566 *    & 0.6604 *     & 0.6792 *     & 0.7151         & 0.7288          \\
			\textit{synthetic-control}    & 0.9677      & 0.9458       & 0.9698       & 0.9567         & 0.9675          \\
			\textit{tic-tac-toe}          & 0.7141      & 0.6589 *     & 0.7109       & 0.7005         & 0.7201          \\
			\textit{titanic}              & 0.7782      & 0.6680 *     & 0.7751       & 0.7822         & 0.7991          \\
			\textit{twonorm}              & 0.9384      & 0.9364       & 0.9388       & 0.9489         & 0.9346          \\
			\textit{wall-following}       & 0.8032      & 0.7964       & 0.8137       & 0.7976         & 0.8199          \\
			\textit{waveform}             & 0.8080 *    & 0.7960 *     & 0.8172 *     & 0.8355         & 0.8317          \\
			\textit{wilt}                 & 0.9472      & 0.9374 *     & 0.9475       & 0.9523         & 0.9538          \\
			\textit{wine}                 & 0.9694      & 0.9625       & 0.9750       & 0.9697         & 0.9622          \\ 
			\textit{Average}              & 0.8146      & 0.8028       & 0.8169       & 0.8317         & 0.8345          \\ \hline
			G/W/L                         & 9/15/35     & 0/2/48       & 8/12/38      & 33 / /         & \textbf{}       \\ 
		\end{longtable}
	\end{ThreePartTable}

	Compared with WNB, CFW, NB, the accuracy of ATFNB on 33 datasets is the highest, which far exceeds WNB (0 datasets), CFW (8 datasets), NB (9 datasets). The average accuracy of ATFNB is 83.17\%, which is significantly higher than those of algorithms, and the improvement of average accuracy is approximately 3\%, 2\%, and 2\%, respectively.
	
	In addition, the average accuracy of CFW-$\beta$ increases by 1.76\%  compared with CFW. This means that the adaptive switching factor can improve the existing two-index NB model. Compared with ATFNB, the average accuracy of CFW-$\beta$ is higher than ATFNB. The reason is that mutual information (class-attribute) and mutual information (attribute-attribute) are included in CFW-$\beta$, which has a more powerful representation than information gain and Pearson correlation coefficient in ATFNB. In Subsection 5.3, models generated by different combinations of indexes are discussed in detail.
	
	Base on the accuracy result, we use a two-tailed t-test at the $p=0.05$ to compare each pair of algorithms beside CFW-$\beta$. Table \ref{T6} summarizes the comparison results on UCI Dataset. From Table \ref{T6}, ATFNB has significant advantages over other weighting algorithms. ATFNB is better than WNB(28 wins and zero loss), CFW(22 wins and 4 loss), and NB(19 wins and 5 loss).
	\begin{table}[htbp]
		\caption{Summary two-tailed t-test results of classification accuracy with regard to ATFNB on UCI dataset}
		\label{T6}
		\begin{tabular*}{1\textwidth}{@{\extracolsep{\fill}} lcccc}
			\toprule
			\textbf{Algorithm} & \textbf{ATFNB}  & \textbf{WNB}   & \textbf{CFW}    & \textbf{NB}    \\
			\midrule
			ATFNB     & ---    & 2(0)  & 12(4)  & 15(5)  \\
			WNB       & 48(28) & ---   & 35(16) & 34(19) \\
			CFW       & 38(22) & 15(5) & ---    & 20(8)  \\
			NB        & 35(19) & 16(9) & 30(11) & ---   \\
			\bottomrule
		\end{tabular*}
		\begin{footnotesize}
			\noindent For each \textit{i(j)}, $i$ represents the number of datasets with higher classification accuracy obtained by the column algorithm than the row algorithm, and $j$ represents the number of datasets in which the column algorithm has a significant advantage over the row algorithm. 
		\end{footnotesize}
	\end{table}\\
	
	Based on the classification accuracy of Table \ref{T5}, we utilize the Wilcoxon signed-rank test to compare four algorithms. Wilcoxon signed-rank test is a non-parametric statistical test, which ranks the performance differences of the two algorithms for each dataset, considering both the sign of the difference and the order of the difference. Tables \ref{T7} shows the ranks calculated by the Wilcoxon test. In Table \ref{T7}, the numbers above the diagonal line indicate the sum of ranks for the datasets of the algorithm in the row that is better than the algorithm in the corresponding column (The sum of the ranks for the positive difference, represented by R+). Each number below the diagonal is the sum of ranks for the datasets in which the algorithm in the column is worse than the algorithm in the corresponding row (The sum of the ranks for the negative difference, represented by R-). According to the critical value table of the Wilcoxon test, for Table \ref{T7}, when $\alpha$=0.05 and n=50, if the smaller of R+ and R- is equal to or less than 434, we consider that two classifiers are significantly different, so we reject the null hypothesis.
	\begin{table}[htbp]
		\caption{Ranks of the Wilcoxon test with regard to ATFNB on UCI dataset}
		\label{T7}
		\begin{tabular*}{1\textwidth}{@{\extracolsep{\fill}} lcccc}
			\toprule
			Algorithm& ATFNB  & WNB  &CFW   & NB   \\
			\midrule
			ATFNB     & ---    & 1268  & 1007.5  & 961.5  \\
			WNB       & 7 & ---   & 308.5 & 391.5 \\
			CFW       & 267.5 & 966.5 & ---    & 771  \\
			NB        & 313.5 & 883.5 & 504 & ---   \\
			\bottomrule
		\end{tabular*}
	\end{table}
	
	\begin{table}[htbp]
		\caption{Summary of the Wilcoxon test with regard to ATFNB on UCI dataset}
		\label{T8}
		\begin{tabular*}{1\textwidth}{@{\extracolsep{\fill}} lcccc}
			\toprule
			Algorithm& ATFNB  & WNB  &CFW   & NB   \\
			\midrule
			ATFNB     & ---    & $\circ$   & $\circ$  & $\circ$\\
			WNB       & $\bullet$  & ---   & $\circ$  & $\circ$  \\
			CFW       &  $\bullet$& $\bullet$ & ---    &   \\
			NB        & $\bullet$ & $\bullet$ & & ---   \\
			\bottomrule
		\end{tabular*}
		\begin{footnotesize}
			\noindent $\bullet$ \quad  indicates that the algorithm in the column is improved compared to the algorithm in the corresponding row.\\
			\noindent $\circ$ \quad indicates that the algorithm in the row is better than the algorithm in the corresponding column.
		\end{footnotesize}
	\end{table}
	According to the results of the Wilcoxon signed rank-sum test, on the UCI dataset, ATFNB is significantly better than WNB ($R^+=1268,R^-=7$), CFW ($R^+=1007.5,R^-=267.5$) and Standard NB ($R^+=961.5,R^-=313.5$).
	
	\subsection{Experimental results on Flavia Dataset}
	In order to further verify the effectiveness of ATFNB, we conduct 15 groups of experiments on Flavia dataset. We randomly divide the data in each group of experiments 30 times and use a two-tailed t-test for the results of 30 experiments. The detailed results are shown in Table \ref{T9}.
	
	\begin{table}[htbp]\centering
		\caption{Classification accuracy comparisons for ATFNB, NB, WNB, CFW, CFW-$\beta$ on Flavia dataset.}
		\label{T9}
		\begin{tabular*}{1\textwidth}{@{\extracolsep{\fill}} llllll}
			\toprule
			\textbf{Group}   & \textbf{NB} & \textbf{WNB} & \textbf{CFW} & \textbf{ATFNB} & \textbf{CFW-$\beta$} \\ \midrule
			\textit{G\_1}    & 0.8253 *    & 0.8506 *     & 0.8552 *     & 0.8805         & 0.9011          \\
			\textit{G\_2}    & 0.8337      & 0.8629       & 0.8742       & 0.8444         & 0.8668          \\
			\textit{G\_3}    & 0.7874 *    & 0.8484       & 0.8312       & 0.8786         & 0.8771          \\
			\textit{G\_4}    & 0.8562 *    & 0.8854 *     & 0.8899       & 0.8987         & 0.9022          \\
			\textit{G\_5}    & 0.8016 *    & 0.8129       & 0.8050 *     & 0.8174         & 0.8177          \\
			\textit{G\_6}    & 0.8822      & 0.8729 *     & 0.8903       & 0.8843         & 0.8801          \\
			\textit{G\_7}    & 0.9134 *    & 0.9137 *     & 0.9322       & 0.9233         & 0.9400          \\
			\textit{G\_8}    & 0.9011      & 0.8812 *     & 0.8927       & 0.8904         & 0.9022          \\
			\textit{G\_9}    & 0.9122      & 0.9100       & 0.9033 *     & 0.9422         & 0.8891          \\
			\textit{G\_10}   & 0.8135 *    & 0.8213 *     & 0.8200 *     & 0.8422         & 0.8399          \\
			\textit{G\_11}   & 0.8572      & 0.8734       & 0.8534       & 0.8799         & 0.8912          \\
			\textit{G\_12}   & 0.8356      & 0.8132 *     & 0.8224       & 0.8233         & 0.8335          \\
			\textit{G\_13}   & 0.7724 *    & 0.7787 *     & 0.7732 *     & 0.7987         & 0.7887          \\
			\textit{G\_14}   & 0.8342 *    & 0.8344       & 0.8322 *     & 0.8458         & 0.8422          \\
			\textit{G\_15}   & 0.9169 *    & 0.9224 *     & 0.9243 *     & 0.9321         & 0.9095          \\ \hline
			\textit{Average} & 0.8495      & 0.8588       & 0.8600       & 0.8721         & 0.8720          \\ \hline
			G/W/L            & 2/2/13      & 0/1/14       & 3/4/11       & 10 / /         & \textbf{}       \\ \bottomrule
		\end{tabular*}
	\end{table}
	
	Comparing ATFNB with other existing classifiers (WNB, CFW, NB), the average accuracy of ATFNB is 87.21\%, which is significantly higher than those of algorithms, and the improvement of average accuracy is approximately 3\%, 1.5\%, 1\%, respectively. In 15 groups of experiments, ATFNB achieved the highest classification accuracy among 10 groups of data, which is far better than NB, WNB, and CFW. 
	
	The average accuracy of CFW-$\beta$ is slightly lower than ATFNB, but the average accuracy of CFW-$\beta$ is higher than CFW. On Flavia, the choice of indexes has little effect on the average accuracy, but adding a switching factor $\beta$ to the model can effectively improve the performance of model.
	
	We Summarize the results of the two-tailed test in Table \ref{T9}, as shown in Table \ref{T10}. In Table \ref{T10}, ATFNB is better than WNB(9 wins and zero loss), CFW(7 wins and 1 loss), and NB(9 wins and zero loss).
	
	\begin{table}[htbp]
		\caption{Summary two-tailed t-test results of classification accuracy with regard to ATFNB on Flavia dataset}
		\label{T10}
		\begin{tabular*}{1\textwidth}{@{\extracolsep{\fill}} lcccc}
			\toprule
			\textbf{Algorithm} & \textbf{ATFNB}  & \textbf{WNB}   & \textbf{CFW}    & \textbf{NB}    \\
			\midrule
			ATFNB     & ---    & 1(0)  & 4(1)  & 2(0)  \\
			WNB       & 14(9) & ---   & 8(3) & 4(1) \\
			CFW       & 11(7) & 7(4) & ---    & 5(1)  \\
			NB        & 13(9) & 11(6) & 10(5) & ---   \\
			\bottomrule
		\end{tabular*}
		\begin{footnotesize}
		\end{footnotesize}
	\end{table}
	
	On the basis of Table \ref{T9}, we use the Wilcoxon signed-rank test to compare four algorithms. According to the critical value table of the Wilcoxon test, for Table \ref{T11}, when $\alpha$=0.05 and n=15, if the smaller of R+ and R- is equal to or less than 25, we consider that two classifiers are significantly different, so we reject the null hypothesis.
	
	\begin{table}[htbp]
		\caption{Ranks of the Wilcoxon test with regard to ATFNB on Flavia dataset}
		\label{T11}
		\begin{tabular*}{1\textwidth}{@{\extracolsep{\fill}} lcccc}
			\toprule
			Algorithm& ATFNB  & WNB  &CFW   & NB   \\
			\midrule
			ATFNB     & ---    & 110  & 96  & 110.5  \\
			WNB       & 10 & ---   & 52 & 89 \\
			CFW       & 24 & 68 & ---    & 87  \\
			NB        & 9.5 & 31 & 33 & ---   \\
			\bottomrule
		\end{tabular*}
	\end{table}
	\begin{table}[htbp]
		\caption{Summary of the Wilcoxon test with regard to ATFNB on Flavia dataset}
		\label{T12}
		\begin{tabular*}{1\textwidth}{@{\extracolsep{\fill}} lcccc}
			\toprule
			Algorithm& ATFNB  & WNB  &CFW   & NB   \\
			\midrule
			ATFNB     & ---    & $\circ$   & $\circ$  & $\circ$ \\
			WNB       & $\bullet$  & ---   &  &   \\
			CFW       &  $\bullet$&  & ---    &   \\
			NB        & $\bullet$ &  & & ---   \\
			\bottomrule
		\end{tabular*}
	\end{table}
	
	In the Flavia dataset, the ATFNB algorithm is compared with WNB ($R^+=110,R^-=10$), CFW ($R^+=96,R^-=24$) and standard NB ($R^+=110.5,R^-=9.5$) has obvious advantages.
	
	\section{Discussion}
	\subsection{The influence of instance and attribute number}
	To further analyze the relationship between the performance of ATFNB and the characteristic of dataset, we observe their performance from two perspectives of instances number and attributes number. In terms of the number of instances, we divide the dataset into two categories: less than 500 instances and greater than or equal to 500 instances. Similar, according to the number of attributes, we divide attributes into two categories: the number of attributes is less than 15, and the number of attributes is greater than or equal to 15. Then, we combine above two criteria and result in four divisions. Finally, we calculate the percentage of the dataset with the highest classification accuracy of ATFNB and competitors (NB, WNB, CFW) in eight divisions. The detailed results are shown in Table \ref{T13}.

	\begin{table}[htbp]\centering
		\caption{ATFNB and competitors obtain the percentage of the dataset with the highest classification accuracy in each division}
		\label{T13}
		\begin{tabular*}{1\textwidth}{@{\extracolsep{\fill}} llccc}
			\toprule
			\multicolumn{2}{c}{Data Characteristics}                                                       & Number & ATFNB (\%)     & Competitors (\%) \\ \midrule
			\multirow{2}{*}{Instance number}     & \textless{}500                                           & 23     & \textbf{78.26} & 21.74            \\
			& $\geq$500                                                     & 27     & \textbf{56.25}          & 43.75            \\ \midrule
			\multirow{2}{*}{Attribute number}    & \textless{}15                                            & 31     & \textbf{67.74}          & 32.26            \\
			& $\geq$15                                                      & 19     & \textbf{63.16}          & 36.84            \\ \midrule
			\multirow{4}{*}{Instance\&Attribute} & \multicolumn{1}{c}{\textless{}500\&\textless{}15}       & 15     & \textbf{73.33} & 26.67            \\
			& \multicolumn{1}{c}{\textless{}500\&$\geq$15}    & 8      & \textbf{87.50} & 12.50            \\
			& \multicolumn{1}{c}{$\geq$500\&\textless{}15}    & 16     & \textbf{62.50} & 37.50            \\
			& \multicolumn{1}{c}{$\geq$500\&$\geq$15} & 11     & 45.45          & \textbf{54.55}   \\ \bottomrule
		\end{tabular*}
	\end{table}

	From Table \ref{T13}, we can clearly find in which circumstance ATFNB performs better than the competitors. Here, we summarize the highlights as follow:\\
	(1) On the datasets with the number of instances less than 500, the percentage of the dataset with the highest classification accuracy of ATFNB (78.26\%) is higher than the number of instances is greater than or equal to 500 (56.25\%).\\
	(2) For datasets with attributes less than 15, the percentage of datasets with the highest classification accuracy of ATFNB (67.74\%) is also higher than that with attributes greater than or equal to 15 (63.16\%).\\
	(3) When the number of instances is less than 500, and the number of attributes is greater than 15, the percentage of the dataset with the highest classification accuracy of ATFNB (87.5\%) is significantly higher than that of the other three types of datasets (73.33\%, 62.50\%, 45.45\%).
	
	The performance of ATFNB has obvious advantages on the datasets whose instance number is smaller than 500, especially attribute number is greater than or equal to 15, such as the dataset “\textit{congressional-voting}”. By contrast, ATFNB does not perform well on datasets with large instances and  attributes. In a word, ATFNB can be perfectly suitable for small data classification, and is not limited by dimensions.

	\subsection{The distribution of the switching factor $\beta$}
	In Section 4.3, we have validated the effectiveness of the switching factor $\beta$ in ATFNB. Here, the distributions of the switching factor $\beta$ in various datasets are further analyzed. We firstly list the interval of the switching factor $\beta$ on 50 UCI datasets as shown in Table \ref{T14}. From Table \ref{T14}, we can summarize that the lower bound of the optimal interval in 11 datasets is greater than 0.5, the upper bound of the optimal interval in 23 datasets is less than 0.5, and the optimal interval of the rest 16 datasets contains 0.5. In ATFNB, the information gain and Pearson correlation coefficient provide different contributions on the 50 UCI datasets. In addition, these results further demonstrates that the switching factor $\beta$ set as a fixed value is unreasonable.
	
	\begin{table}[htbp]
		\caption{The interval of switching factor $\beta$ on 50 UCI Datasets}
		\label{T14}
		\begin{tabular}{@{}llc|llc@{}}
			\toprule
			Dataset                       & Interval ($\beta$)         & Mark & Dataset                    & Interval ($\beta$)         & Mark \\ \midrule
			\textit{abalone}              & {[}0.7122, 0.8311{]} & $\bigcirc$  & \textit{libras}            & {[}0.5377, 0.8832{]} & $\bigcirc$    \\
			\textit{acute}                & {[}0.4418, 0.9433{]} & $\bigtriangleup$    & \textit{low-res-spect}     & {[}0.6552, 0.7211{]} & $\bigcirc$    \\
			\textit{aggregation}          & {[}0.5529, 0.8832{]} & $\bigtriangleup$    & \textit{lymphography}      & {[}0.3344, 0.4834{]} & $\square$    \\
			\textit{balance-scale}        & {[}0.3233, 0.4537{]} & $\square$    & \textit{magic}             & {[}0.6733, 0.8122{]} & $\bigcirc$    \\
			\textit{bank}                 & {[}0.4198, 0.7691{]} & $\bigtriangleup$    & \textit{mammographic}      & {[}0.1229, 0.3879{]} & $\square$    \\
			\textit{banknote}             & {[}0.3144, 0.3914{]} & $\square$    & \textit{promoters}         & {[}0.4876, 0.8867{]} & $\bigtriangleup$    \\
			\textit{blood}                & {[}0.2243, 0.5532{]} & $\bigtriangleup$    & \textit{splice}            & {[}0.0512, 0.1321{]} & $\square$    \\
			\textit{breast-cancer}        & {[}0.2311, 0.3521{]} & $\square$    & \textit{nursery}           & {[}0.5211, 0.5908{]} & $\bigcirc$    \\
			\textit{breast-tissue}        & {[}0.3566, 0.4513{]} & $\square$    & \textit{page-blocks}       & {[}0.6322, 0.7109{]} & $\bigcirc$    \\
			\textit{bupa}                 & {[}0.1533, 0.6588{]} & $\bigtriangleup$    & \textit{pima}              & {[}0.1566, 0.3118{]} & $\square$    \\
			\textit{car}                  & {[}0.3211, 0.3987{]} & $\square$    & \textit{planning}          & {[}0.1829, 0.4721{]} & $\square$   \\
			\textit{chart\_Input}         & {[}0.4592, 0.8311{]} & $\bigtriangleup$    & \textit{post-operative}    & {[}0.0187, 0.2100{]} & $\square$    \\
			\textit{climate-simulation}   & {[}0.2301, 0.3255{]} & $\square$    & \textit{robotnavigation}   & {[}0.7122, 0.7830{]} & $\bigcirc$    \\
			\textit{congressional-voting} & {[}0.2199, 0.3472{]} & $\square$    & \textit{seeds}             & {[}0.4288, 0.8543{]} &$\bigtriangleup$    \\
			\textit{connectionist}        & {[}0.0912, 0.1388{]} & $\square$    & \textit{sonar}             & {[}0.0521, 0.1487{]} & $\square$    \\
			\textit{dermatology}          & {[}0.3365, 0.8987{]} & $\bigtriangleup$    & \textit{soybean}           & {[}0.2759, 0.3108{]} & $\square$    \\
			\textit{diabetes}             & {[}0.2355, 0.5243{]} & $\bigtriangleup$    & \textit{spect}             & {[}0.1802, 0.2499{]} & $\square$    \\
			\textit{Ecoli}                & {[}0.7360, 0.9211{]} & $\bigcirc$    & \textit{synthetic-control} & {[}0.2480, 0.4033{]} & $\square$    \\
			\textit{energy-y1}            & {[}0.3211, 0.9219{]} & $\bigtriangleup$    & \textit{tic-tac-toe}       & {[}0.1213, 0.1870{]} & $\square$    \\
			\textit{fertility}            & {[}0.0511, 0.4390{]} & $\square$    & \textit{titanic}           & {[}0.4697, 0.6122{]} & $\bigtriangleup$    \\
			\textit{glass}                & {[}0.1229, 0.1833{]} & $\square$    & \textit{twonorm}           & {[}0.5833, 0.6291{]} & $\bigcirc$    \\
			\textit{haberman-survival}    & {[}0.2166, 0.6345{]} & $\bigtriangleup$    & \textit{wall-following}    & {[}0.4128, 0.4736{]} & $\square$    \\
			\textit{iris}                 & {[}0.3522, 0.8799{]} & $\bigtriangleup$    & \textit{waveform}          & {[}0.7398, 0.7933{]} & $\bigcirc$    \\
			\textit{jain}                 & {[}0.3409, 0.8577{]} & $\bigtriangleup$    & \textit{wilt}              & {[}0.6103, 0.6899{]} & $\bigcirc$    \\
			\textit{knowledge}            & {[}0.3012, 0.4522{]} & $\square$    & \textit{wine}              & {[}0.3881, 0.9220{]} & $\bigtriangleup$    \\ \bottomrule
		\end{tabular}
		
		\begin{footnotesize}
			\noindent $\bigcirc$ \quad  indicates that the lower bound value of $\beta$ interval is greater than 0.5.\\
			\noindent $\square$ \quad  indicates that the upper bound value of $\beta$ interval is less than 0.5.\\
			\noindent $\bigtriangleup$  \quad   indicates 0.5 is in the interval.\\
		\end{footnotesize}
	\end{table}

	\begin{table}[htbp]\centering
		\caption{The relationship between the switching factor $\beta$ in the ATFNB and the characteristics of the dataset}
		\label{T15}
		\begin{tabular*}{1\textwidth}{@{\extracolsep{\fill}} llcccc}
			\toprule
			\multicolumn{2}{l}{Data Characteristics}                                                    & Number & $\bigcirc$ (\%)  & $\bigtriangleup$  (\%)&  $\square$ (\%) \\ \hline
			\multicolumn{1}{c}{\multirow{2}{*}{Instance number}}  & \textless{}500                                  & 23     & 8.70      & 39.13     & 52.17     \\
			\multicolumn{1}{c}{}                                  & $\geq$500                                  & 27     & 33.33     & 25.93     & 40.74    \\ \hline
			\multicolumn{1}{c}{\multirow{2}{*}{Attribute number}} & \textless{}15                                     & 31     & 19.35     & 38.71     & 41.94     \\
			\multicolumn{1}{c}{}                                  & $\geq$15                                   & 19     & 26.32     & 21.05     & 52.63     \\ \hline
			\multirow{4}{*}{Instance\&Attribute}                  & \textless{}500\&\textless{}15       & 15     & 6.66      & 46.67     & 46.67     \\
			& \textless{}500\&$\geq$15    & 8      & 12.50     & 25.00     & 62.50     \\
			& $\geq$500\&\textless{}15    & 16     & 31.25     & 31.25     & 37.50     \\
			& $\geq$500\&$\geq$15 & 11     & 36.36     & 18.18     & 45.46     \\ \bottomrule
		\end{tabular*}
	\end{table}
	To further investigate the relationship between the distribution of switching factor $\beta$ and the characteristics of dataset, we apply the same division criteria as Section 5.1 on 50 UCI datasets and summarize detailed results in Table \ref{T15}.
	From Table \ref{T15}, we can observe the preference between the data characteristic and  the distribution of the switching factor $\beta$, and summarize the highlights as follows:  \\
	(1) If the number of instance is less than 500, the upper bound value of $\beta$ in 52.17\% of the datasets is less than 0.5. The number of datasets is more than 500, and the upper bound value of $\beta$ in 40.74\% of the datasets is less than 0.5. \\
	(2) From the perspective of the number of attributes, regardless of the number of attributes, the upper bound value of $\beta$ is less than 0.5 in most datasets.\\
	(3) Considering the number of instance and attributes simultaneously, the upper bound value of $\beta$ in 62.50\% of the datasets with instances less than 500 and attributes greater than 15 is less than 0.5. On the dataset with instances greater than 500 and attribute number greater than 15, the upper bound value of $\beta$ in 45.46\% of the datasets is less than 0.5. 
	
	Based on the results in Table \ref{T15}, the upper bound value of $\beta$ is less than 0.5 in most datasets. We can conclude that ATFNB pays attention to Pearson correlation coefficient between attributes, especially in small instances and high-dimensional datasets. 
	
	\subsection{The impact of different index combinations}
	
\begin{figure}[!htbp]\centering
		\includegraphics[width=0.95\textwidth]{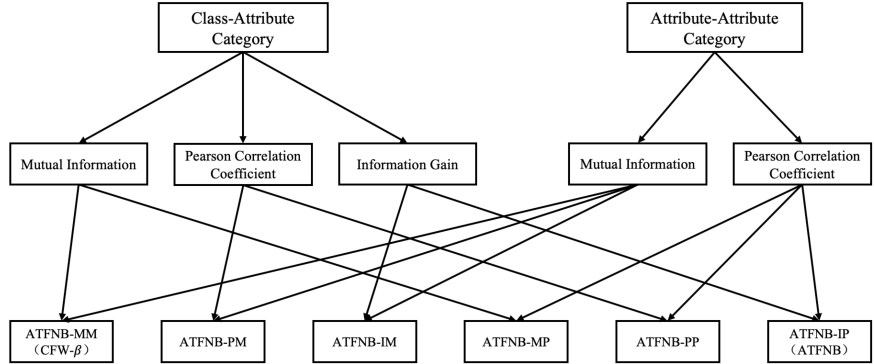}
		\caption{The index selection of each combination}
		\label{F5.3.1}
	\end{figure}

	The ATFNB framework contains two categories, and each category provides several popular indexes to represent the characteristic of datasets. Now, in order to analyze the impact of different index combinations, we select two any indexes from two categories respectively. Excluding the gain ratio from class-attribute category, six weighted NB models can be constructed as shown in Figure \ref{F5.3.1}. Notably, ATFNB-IP and ATFNB-MM are equal to ATFNB and CFW-$\beta$ respectively.

	\begin{figure}[!htbp]\centering
		\includegraphics[width=0.95\textwidth]{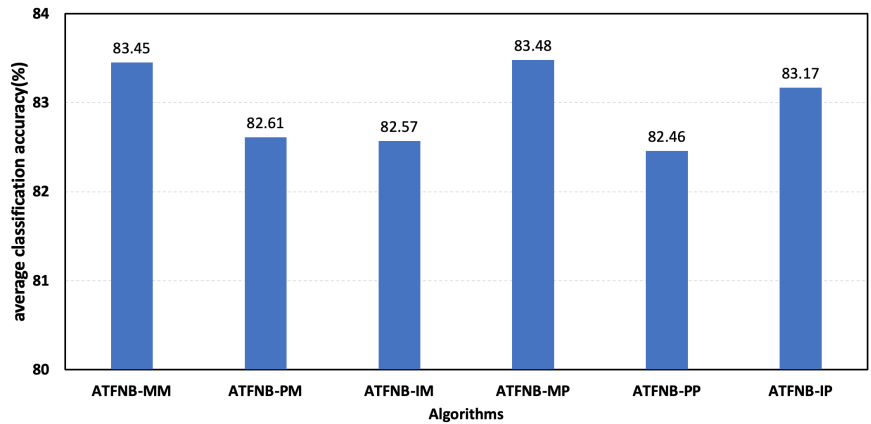}
		\caption{Average accuracy of six combinations}
		\label{F5.3.2}
	\end{figure}
	Then, we compare six combinations on the 50 UCI datasets, and the average accuracy of six combinations are shown in Figure \ref{F5.3.2}. From Figure \ref{F5.3.2}, it can be seen that ATFNB-PP receives the lowest, but average accuracy of ATFNB-PP outperforms the basic NB and NB(0.8146), WNB(0.8028) and CFW(0.8169). This further demonstrates the effectiveness of ATFNB framework with adaptive switching factor. In addition, compare with three indexed from class-attribute category, the average accuracy of ATFNB-M*(denotates ATFNB-MP and ATFNB-MM) is better than ATFNB-P* and ATFNB-I*. This means mutual information from class-attribute category is more signification than Pearson correlation coefficient and Information gain.
	\section{Conclusions and future work}
	In this paper, we propose a general framework for adaptive Two-index Fusion attribute weighted NB(ATFNB) to overcome the problems of the existing weighted methods, such as the poor representation ability with single index and the fusion problem of two indexes. ATFNB can select any one index from attribute-attribute category and class-attribute category, respectively. Then, switching factor $\beta$ is introduced to fuse two indexes and inferred by a quick algorithm. Finally, the weight of each attribute is calculated using the optimal value $\beta$ and integrated into NB classifier to improve the accuracy. The experimental results on 50 benchmark datasets and a Flavia dataset show that ATFNB outperforms the basic NB and state-of-the-art filter weighted NB models. In addition, we incorporate the switching factor $\beta$ into CFW. The results demonstrate the improved model CFW-$\beta$ significantly increase accuracy compared to CFW without the adaptive switching factor $\beta$. 
	
	In the future work, there are two direction to further improve NB model. Firstly, ATFNB maybe consider more than two indexes from different data description categories. Secondly, we hope design more new indexes to represent the correlation between class-attribute or attribute-attribute.

	\bibliography{mybibfile}
	
\end{document}